# Experimental Identification of Hard Data Sets for Classification and Feature Selection Methods with Insights on Method Selection

Cuiju Luan[a], [*], Guozhu Dong[b]

[a] *College of Information Engineering, Shanghai Maritime University, Shanghai, 201306, China*
[b] *Department of Computer Science and Engineering, Wright State University, Dayton, Ohio 45435, USA*

**ABSTRACT**

The paper reports an experimentally identified list of benchmark data sets that are hard for representative classification and feature selection methods. This was done after systematically evaluating a total of 48 combinations of methods, involving eight state-of-the-art classification algorithms and six commonly used feature selection methods, on 129 data sets from the UCI repository (some data sets with known high classification accuracy were excluded). In this paper, a data set for classification is called hard if none of the 48 combinations can achieve an AUC over 0.8 and none of them can achieve an F-Measure value over 0.8; it is called easy otherwise. A total of **15** out of the 129 data sets were found to be hard in that sense. This paper also compares the performance of different methods, and it produces rankings of classification methods, separately on the hard data sets and on the easy data sets. This paper is the first to rank methods separately for hard data sets and for easy data sets. It turns out that the classifier rankings resulting from our experiments are somehow different from those in the literature and hence they offer new insights on method selection. It should be noted that the Random Forest method remains to be the best in all groups of experiments.

## 1 Introduction

When faced with a classification job, an analyst will often want to select the best methods for the application; this can be a daunting task since there are a large number of methods available. Users will need insights, such as rankings of the methods, to guide them to make the best selection, and to go through the selection process in an easy-to-handle manner. Several studies on the experimental evaluation of various methods for classification have been reported recently [1, 2]. Reference [1] is a main representative of such studies, which used 121 data sets to evaluate 179 classifiers.

---

[*] Corresponding author.
  *E-mail addresses*: cjluan@shmtu.edu.cn (C. Luan), guozhu.dong@wright.edu (G. Dong).

However, to the best of our knowledge, all previous studies evaluated and ranked classification methods by considering all data sets in one pool – they did not distinguish the data sets based on their hardness. Moreover, there was no systematic study to identify which classification benchmark data sets are hard for traditional classification methods, and there were no rankings of methods based on their performance on hard data sets only. Filling these gaps is important, as the identified hard data sets can help future studies to develop new classification algorithms to complement existing classification algorithms, and the ranking of methods on hard data sets can help users select the best method when they are working with potentially hard data sets. We plan to fill this gap in this study.

This study will evaluate both classification algorithms and feature selection methods in combination. Specifically, it will identify hard data sets for which no combinations of representative classification algorithms and feature selection methods can produce accurate classification models. Moreover, the study will use the area under the ROC (AUC) and F-Measure, instead of the accuracy measure [1], to evaluate the performance of classification models. These measures were chosen based on the recent consensus that the accuracy measure has significant shortcomings when compared with the above two measures, especially AUC.

To identify a list of benchmark data sets that are hard for representative classification and feature selection methods, we perform a systematical evaluation of 48 combinations, involving eight representative classification algorithms and six commonly used feature selection methods, on 129 data sets from the UCI repository. We note that some data sets with known high classification accuracy based on results of Fernández-Delgado et al. [1] were excluded in our experiments.

For ease of discussion, a data set for classification will be called hard if none of the 48 combinations can achieve an AUC over 0.8 and none of the 48 combinations can achieve an F-Measure value over 0.8; it is called easy otherwise. A total of 15 out of the 129 data sets were found to be hard in our experiments.

This paper also compares the performance of different methods separately on the hard data sets and on the easy data sets. This was done based on their performance on data sets for which complete results were obtained for all of the 48 combinations. It turns out that the method rankings resulting from our experiments are somehow different from those in the literature and hence they offer new insights on method selection.

The rest of the paper is organized as follows. Section 2 describes the classification algorithms and feature selection methods used in this study. Section 3 describes the data sets included in this study. Section 4 gives the experiment settings and the evaluation measures used. Section 5 presents the experimental results and the associated analysis. Section 6 concludes the paper.

## 2 Algorithms Used in the Study

In the experiments, we used multiple commonly-used representative classification algorithms and feature selection methods. The classification algorithms we used are Boosting, Decision Tree, Random Forest, Nearest Neighbor, Logistic Regression, and Support Vector Machine (SVM). The feature selection methods we used are correlation based method, information gain based method, and the relief-f method (all of which are filter based methods).

During the experiments, we also considered the wrapper based method, but we decided to exclude it due to its computational expensiveness (see Table X in Appendix) (it is seldom used in practice [3] due to the same reason).

All the classification algorithms and feature selection methods we used are as implemented in Weka 3.8.0 [4]. More details are given in the next two subsections.

**2.1 Classification Algorithms and Parameter Settings**

We selected representative classification algorithms, partly based on several papers that reported systematic evaluation of classification algorithms and partly based on common knowledge. In particular, reference [1] showed Random Forest and SVM are often better than the others, and reference [5] gave a list of common-used successful classification algorithms. We selected Boosting, Decision Tree, Random Forest, Nearest Neighbor, Logistic Regression and SVM, as the representatives of existing classification algorithms. Table 1 shows the correspondence of classification algorithms and their implementations in Weka that we used in our experiments. Some of the classification algorithms given in Table 1 have multiple versions due to different parameter settings, yielding a total of eight classification algorithms (discussed below).

**Table 1**
Classification algorithms and their implementations in Weka.

| Classification algorithm | Classifier in Weka | Abbreviation |
|---|---|---|
| Boosting | AdaBoostM1 | AdaBoost |
| Decision Tree | J48 | J48 |
| Random Forest | RandomForest | RF |
| K Nearest Neighbor | IBk | IBk |
| Logistic | Logistic | LOG |
| SVM | LibSVM | SVM |

To better evaluate the six algorithms, *two J48 and two SVM classifiers* were examined. For J48, we examined the classifier using the default parameter values, and the other using no pruning and using Laplace smoothing (J48(-U-A)). The two LibSVM classifiers are polynomial (SVM-PN) and radial basis function (SVM-RBF) kernels respectively. For IBk, we used K=10 and crossValidate=True, with which the system will find the optional K value between 1 and 10. More details are given below.

1) AdaBoost uses the M1 method [6] with DecisionStump as base classifiers.
2) J48 implements a pruned C4.5 [7] decision tree algorithm, with parameters confidenceFactor=0.25 and minNumObj=2.
3) J48(-U-A) implements a unpruned C4.5 decision tree algorithm with Laplace-based smoothening.
4) RF builds a forest of random trees [8], with parameters bagSizePercent=100 and unlimited depth.
5) IBk implements the K-nearest neighbors classifier [9], selecting the optimal value of K between 1 and 10 based on cross-validation.

6) LOG builds and uses a multinomial logistic regression model with a ridge estimator [10], with parameter ridge=1.0E-8.
7) SVM-PN uses the LibSVM library with polynomial kernel, with parameters SVMType=C-SVC, cost ={0.25, 0.5, 1}, degree={1, 2, 3}, gamma={0.110,0.01,0.1} and coef0={0, 1}.
8) SVM-RBF uses the LibSVM library with radial basis function kernel, with parameters SVMType=C-SVC, cost={$2^{-5}$, $2^{-3}$, …, $2^{15}$} and gamma={$2^{-15}$, $2^{-13}$, …, $2^{3}$}.

## 2.2 Feature Selection Methods

Feature selection methods are used to remove irrelevant, redundant, or noisy attributes, with the aim of speeding up the computation and improving the accuracy [3, 12, 13]. Our experiments examine the effectiveness of feature selection methods when used with various classification algorithms.

Many feature selection methods have been proposed, each with its own pros and cons. We selected the following filter methods because they are commonly used [12, 14]: the correlation based method, the information gain based method and the relief-f method. We used Weka's implementation of the three feature selection methods.

In order to better evaluate these methods, we used two different parameter settings for Information Gain and Relief-F, which dictate how many features are selected. For ease of discussion, we consider "no attribute selection" as a feature selection method. Therefore, a total of 6 feature selection methods are considered. We now give some more details on the feature selection methods.

1) CFS uses CfsSubsetEval as the attribute evaluator to evaluate the worth of a subset of attributes by considering the individual predictive ability of each feature along with the degree of redundancy between them, and uses BestFirst as the search method to searches the space of attribute subsets by greedy hill-climbing augmented with a backtracking facility.
2) IG1 uses InfoGainAttributeEval as the attribute evaluator to evaluate the worth of attributes by measuring the information gain with respect to the class, and uses Ranker as the search method. If the total number of features is no more than 50, IG1 selects 80 percent of the features, and IG1 selects 40 features otherwise.
3) IG2 differs from IG1 as follows. It selects 60 percent of the features if the total number of features is no more than 50, and it selects 25 features otherwise.
4) RLF1 uses ReliefAttributeEval as the attribute evaluator to evaluate the worth of attributes by repeatedly sampling an instance and considering the value of the given attribute for the nearest instance belonging to the same and different classes. It uses Ranker as the search method. It selects 60 percent of the features if the total number of features is no more than 50, and it selects 25 features otherwise.
5) RLF2 differs from RLF1 as follows. It selects 60 percent of the features if the total number of features is no more than 50, and it selects 25 features otherwise.
6) NO means "no attribute selection is performed".

## 3 Data Sets Included in the Study

Our experiments used 129 data sets, all from the UCI repository [15]. Table 2 lists the 98 data sets for which complete results for all of the 48 combinations were obtained; the remaining data sets are in Table XI of the Appendix.

**Table 2**
Details of 98 data sets.

| Data set | #Instance | #Attribute | Data set | #Instance | #Attribute |
|---|---|---|---|---|---|
| abalone | 4177 | 9 | heart | 270 | 14 |
| anneal | 798 | 39 | heart-cleveland | 303 | 14 |
| arrhythmia | 452 | 263 | heart-switzerland | 123 | 13 |
| australian | 690 | 15 | heart-va | 200 | 13 |
| balloons_a | 20 | 5 | hepatitis | 155 | 20 |
| balloons_b | 20 | 5 | hill-valley | 1212 | 101 |
| balloons_c | 20 | 5 | leaf | 340 | 16 |
| balloons_d | 16 | 5 | led-display | 1000 | 8 |
| bankrupt_qualitative | 250 | 7 | letter-recognition | 13339 | 17 |
| biodeg | 1055 | 42 | lung-cancer | 32 | 57 |
| blogger | 100 | 6 | mfeat | 2000 | 650 |
| breast-cancer | 286 | 10 | monks-1 | 556 | 7 |
| breast-cancer-wisc | 699 | 10 | monks-2 | 601 | 7 |
| breast-cancer-wisc-diag | 569 | 31 | monks-3 | 554 | 7 |
| breast-cancer-wisc-prog | 198 | 34 | occupancy | 20560 | 7 |
| breast-tissue | 106 | 10 | phishing | 11055 | 31 |
| chronic_kidney_disease | 400 | 25 | pima | 768 | 9 |
| climate | 540 | 21 | pittsburg-bridges-material | 106 | 8 |
| congressional-voting | 435 | 17 | pittsburg-bridges-rel-l | 103 | 8 |
| contrac | 1473 | 10 | pittsburg-bridges-span | 92 | 8 |
| cortex_nuclear | 1080 | 82 | pittsburg-bridges-t-or-d | 102 | 8 |
| credit_card | 30000 | 24 | pittsburg-bridges-type | 105 | 8 |
| crx | 690 | 16 | planning | 182 | 13 |
| data_banknote_authentication | 1372 | 5 | post-operative | 90 | 9 |
| dr | 1151 | 20 | primary-tumor | 330 | 18 |
| dresses_attribute_sales | 500 | 14 | seismic-bumps | 2584 | 19 |
| eeg_data | 14980 | 15 | shuttle | 58000 | 10 |
| electricity-board | 45781 | 5 | spect | 265 | 23 |
| fertility_diagnosis | 100 | 10 | statlog-australian-credit | 690 | 15 |
| flags | 194 | 29 | statlog-german-credit | 1000 | 25 |
| foresttypes | 523 | 28 | statlog-heart | 270 | 14 |
| gesture_phase_a1_raw | 1747 | 20 | statlog-image | 2310 | 19 |
| gesture_phase_a1_va3 | 1743 | 33 | statlog-landsat | 6435 | 37 |
| gesture_phase_a2_raw | 1264 | 20 | statlog-shuttle | 58000 | 10 |
| gesture_phase_a2_va3 | 1260 | 33 | statlog-vehicle | 846 | 19 |
| gesture_phase_a3_raw | 1834 | 20 | student-mat | 395 | 33 |
| gesture_phase_a3_va3 | 1830 | 33 | student-por | 649 | 33 |
| gesture_phase_b1_raw | 1073 | 20 | teaching | 151 | 6 |
| gesture_phase_b1_va3 | 1069 | 33 | thoracic_surgery_data | 470 | 17 |
| gesture_phase_b3_raw | 1424 | 20 | titanic | 2201 | 4 |
| gesture_phase_b3_va3 | 1420 | 33 | urban_land_cover | 675 | 148 |
| gesture_phase_c1_raw | 1111 | 20 | user_modeling | 403 | 6 |
| gesture_phase_c1_va3 | 1107 | 33 | vehicle | 846 | 19 |
| gesture_phase_c3_raw | 1448 | 20 | wholesale customers data_new | 440 | 8 |
| gesture_phase_c3_va3 | 1444 | 33 | wilt | 4839 | 6 |
| gesture_phase_raw | 9901 | 20 | wine | 178 | 14 |
| gesture_phase_va3 | 9873 | 33 | wine-quality-red | 1599 | 12 |
| glass | 214 | 10 | wine-quality-white | 4898 | 12 |
| haberman-survival | 306 | 4 | yeast | 1484 | 9 |

Below we discuss where the data sets are from and how we selected them.

(a) From the data sets studied in [1], we first selected 34 data sets such that the maximum reported accuracy of [1] is below 0.8. (This saved our effort by eliminating the data sets having high known accuracy.) Then we added another 11 data sets that are variations (sharing the same data set name at UCI) of some of the 34 data sets. As a result, this group has a total of *45* data sets.

(b) Because [1] only dealt with UCI data sets dated before March 2013, we examined the 91 UCI data sets whose dates are between January 2013 and June 2016. From these we selected 32 and excluded the other 59 for reasons such as "too many instances", "having no classification attribute", "having complex data structure requiring preprocessing", "having no data", and "inaccessible". Some data sources provide multiple versions (e.g. 15 for actrecog, 16 for gesture, 10 for mhealth, 2 for student); we took each version as a different data set (e.g. 15 data sets from the 15 versions of actrecog). This group has a total of *71* data sets.

(c) We also examined data sets having no dates marked at UCI, from which 10 data sets are included (the others are excluded due to complex data). Among the data sets, ballons has 4 versions, yielding extra data sets. So this group has a total of *13* data sets.

Among the 129 data sets from the three groups, there were 31 (listed in Table XI in Appendix) for which experiments could not be completed, hence they were excluded from Table 2. One of the 31 came from the (a) group and the other 30 from the (b) group.

## 4 Experimental Settings and Evaluation Measures

We used 10-fold cross validation to evaluate classification performance. For each fold of each data set, a classification model is built from the other 9 folds, using each of the 48 combinations involving eight classifiers and six feature selection methods.

As widely noted in the literature, the simple accuracy measure may be not adequate for imbalanced data sets. As some data sets used in our experiments are not balanced, we did not use the accuracy measure; we used the AUC and F-Measure measures instead.

AUC is equivalent to the probability that the underlying classifier will rank a randomly chosen positive instance higher than a randomly chose negative instance [16]. It is also called ROC Area in Weka. The F-Measure is the harmonic mean of Precision and Recall. AUC has several desirable properties as a classification performance measure, such as being decision threshold independent and invariant to a priori class probabilities. AUC is widely accepted as one of the best ways to evaluate a classifier's performance [17] and it has been widely used to measure the performance in classification. We chose to include the F-Measure, in order to complement the AUC, and also to indicate the strength of the classifier in terms of Precision and Recall. (Reference [18] pointed out that for some situations, namely when the ROC curves cross, AUC has some weakness and it may give potentially misleading information.)

We note that, normally, AUC and F-measure are only defined for two-class problems. This paper also considers multi-class problems (having more than 2 classes). For such problems, the measure values are computed by weighted average of a number of two class problems (one for each class, defined as the class vs the union of the other classes), as is done by weka.

## 5 Experimental Results and Discussion

This section first presents the **15** identified hard data sets. It then analyzes the performance of classification and feature selection methods under several different conditions. It presents, for each hard data set, the combinations that are the best or worst for the data set. Based on that, it identifies the most frequent best classification algorithm, the most frequent best feature selection method, and so on. Similarly, it presents the worst combinations and identifies the most frequent worst methods. Finally, it also gives rankings of classification algorithms based on the number of data sets where they are the best and base on the average AUC; the rankings are given separately for the hard data sets and for the easy data sets. It should be noted that the rankings are based solely on results on the 98 data sets for which complete results were obtained for all of the 48 combinations.

For ease of discussion, we introduce a few terms and notations. For each data set, let *max48AUC* denote the highest AUC achieved by the 48 classification-algorithm and feature-selection-method combinations, and similarly let *max48FMeasure* denote the highest F-Measure. We say a data set is a *hard data set* if both max48AUC and max48FMeasure are no higher than 0.8, and we call the other data sets as *easy data sets*.

As the detailed experiment results require too much space, they are not listed here; they can be found as supplementary materials at http://cecs.wright.edu/~gdong/harddata/.

### 5.1 The 15 Hard Data Sets We Identified

Table 3 lists the data sets for which max48AUC is less than or equal to 0.8, and Table 4 lists the data sets for which max48FMeasure is less than or equal to 0.8. In both tables, the data sets are listed in increasing measure value order.

**Table 3**
Data sets having max48AUC <= 0.8.

| Data set | Max48AUC | Data set | Max48AUC |
|---|---|---|---|
| post-operative | 0.536 | fertility_diagnosis | 0.707 |

| planning | 0.588 | breast-cancer | 0.711 |
|---|---|---|---|
| dresses_attribute_sales | 0.59 | titanic | 0.755 |
| statlog-australian-credit | 0.61 | spect | 0.758 |
| heart-switzerland | 0.626 | seismic-bumps | 0.764 |
| heart-va | 0.632 | credit_card | 0.767 |
| congressional-voting | 0.632 | breast-cancer-wisc-prog | 0.771 |
| thoracic_surgery_data | 0.673 | statlog-german-credit | 0.797 |
| haberman-survival | 0.691 | primary-tumor | 0.799 |
| contrac | 0.704 | pittsburg-bridges-span | 0.8 |

**Table 4**

Data sets having max48FMeasure <= 0.8.

| Data set | Max48FMeasure | Data set | Max48FMeasure |
|---|---|---|---|
| heart-va | 0.399 | wine-quality-red | 0.694 |
| heart-switzerland | 0.434 | teaching | 0.702 |
| primary-tumor | 0.44 | arrhythmia | 0.709 |
| student-por | 0.443 | hill-valley | 0.715 |
| student-mat | 0.455 | haberman-survival | 0.716 |
| congressional-voting | 0.558 | pittsburg-bridges-span | 0.716 |
| contrac | 0.564 | breast-cancer | 0.735 |
| heart-cleveland | 0.592 | led-display | 0.741 |
| dresses_attribute_sales | 0.602 | spect | 0.743 |
| yeast | 0.608 | gesture_phase_c1_va3 | 0.749 |
| pittsburg-bridges-type | 0.625 | gesture_phase_a1_va3 | 0.75 |
| post-operative | 0.626 | breast-tissue | 0.756 |
| electricity-board | 0.659 | pittsburg-bridges-rel-l | 0.759 |
| abalone | 0.666 | dr | 0.77 |
| statlog-australian-credit | 0.668 | titanic | 0.771 |
| gesture_phase_c3_va3 | 0.675 | statlog-german-credit | 0.774 |
| gesture_phase_a2_va3 | 0.676 | pima | 0.777 |
| planning | 0.68 | lung-cancer | 0.779 |
| gesture_phase_va3 | 0.682 | leaf | 0.783 |
| flags | 0.689 | gesture_phase_b3_va3 | 0.797 |
| wine-quality-white | 0.692 | | |

Table 5 reports the hard data sets having both max48AUC and max48FMeasure less than or equal to 0.8 --- they are precisely those that appear in both Tables 3 and 4.

**Table 5**

Hard data sets (having max48AUC <= 0.8 & max48FMeasure <= 0.8).

| Data set | Max48AUC | Max48FMeasure |
|---|---|---|
| post-operative | 0.536 | 0.626 |
| planning | 0.588 | 0.68 |
| dresses_attribute_sales | 0.59 | 0.602 |
| statlog-australian-credit | 0.61 | 0.668 |
| heart-switzerland | 0.626 | 0.434 |
| heart-va | 0.632 | 0.399 |
| congressional-voting | 0.632 | 0.558 |
| haberman-survival | 0.691 | 0.716 |
| contrac | 0.704 | 0.564 |
| breast-cancer | 0.711 | 0.735 |
| titanic | 0.755 | 0.771 |
| spect | 0.758 | 0.743 |
| statlog-german-credit | 0.797 | 0.774 |
| primary-tumor | 0.799 | 0.44 |
| pittsburg-bridges-span | 0.8 | 0.716 |

**5.2 Best and Worst Method Combinations for the Hard Data Sets**

For each hard data set, we identified the best combination of classification and feature selection methods obtaining the highest AUC; the result is reported in Table 6. Summarizing the table we have the following:

- At the individual algorithm level, RF & IG1 and RF & NO make the most-frequent best combinations (being the best for 3 hard data sets).
- The best combinations for 10 (66.7%) of the 15 hard data sets involve the use of feature selection methods; the best combinations for 2 (13.3%) do not use feature selection methods; for the remaining 3 (20%) data sets, multiple combinations achieved the best AUC, some of which involve feature selection methods and some do not.
- Focusing on the feature selection methods, we see that IG1, IG2 and NO (no attribute selection) are used in the best combinations for 5 hard data sets (33.3%), RLF1 is used for 3 hard data sets (20%), CFS is used for 2 hard data sets (13.3%), RLF2 is used for 1 hard data sets (6.7%).
- Focusing on classification algorithms, we see that RF appears in the best combinations of 7 hard data sets (46.7%), LOG appears in 4 (26.7%), AdaBoost and SVM-RBF each appears in 2 (13.3%).

So, for the 15 hard data sets, RF is the most-frequent best classification algorithm, IG, and NO (no attribute selection) are the most-frequent best feature selection methods, and the combinations of RF & IG1 and RF & NO are the most-frequent best combinations.

**Table 6**

Classification & feature selection methods giving best AUC for hard data sets.

| Data set | #Instance | #Attribute | #SelAttr | Classifier | FS Method | Max48AUC |
|---|---|---|---|---|---|---|
| post-operative | 90 | 9 | 7 | SVM-RBF | CFS | 0.536 |
| planning | 182 | 13 | 9 | SVM-RBF | IG2 | 0.588 |
| dresses_attribute_sales | 500 | 14 | 9 | AdaBoost | IG2 | 0.59 |
| statlog-australian-credit | 690 | 15 | 10 | RF | IG2 | 0.61 |

| Data set | #Instance | #Attribute | #SelAttr | Classifier | FS method | Min48AUC |
|---|---|---|---|---|---|---|
| heart-switzerland | 123 | 13 | 11 | RF | IG1 | 0.626 |
| heart-va | 200 | 13 | 13 | RF | NO | 0.632 |
| congressional-voting | 435 | 17 | 11 | LOG | IG2 | 0.632 |
| haberman-survival | 306 | 4 | 4 | LOG | NO | 0.691 |
| haberman-survival | 306 | 4 | 4 | LOG | IG1 | 0.691 |
| haberman-survival | 306 | 4 | 4 | LOG | RLF1 | 0.691 |
| contrac | 1473 | 10 | 9 | RF | RLF1 | 0.704 |
| breast-cancer | 286 | 10 | 9 | AdaBoost | IG1 | 0.711 |
| titanic | 2201 | 4 | 4 | RF | NO | 0.755 |
| titanic | 2201 | 4 | 4 | RF | IG1 | 0.755 |
| titanic | 2201 | 4 | 4 | RF | RLF1 | 0.755 |
| spect | 265 | 23 | 15 | LOG | IG2 | 0.758 |
| statlog-german-credit | 1000 | 25 | 21 | RF | IG1 | 0.797 |
| statlog-german-credit | 1000 | 25 | 16 | RF | RLF2 | 0.797 |
| primary-tumor | 330 | 18 | 18 | RF | NO | 0.799 |
| pittsburg-bridges-span | 92 | 8 | 8 | LOG | NO | 0.8 |
| pittsburg-bridges-span | 92 | 8 | 5 | LOG | CFS | 0.8 |

We now turn to the worst combination of classification and feature selection methods getting lowest AUC for the hard data sets; the result is given in Table 7. Summarizing the table we have the following:

- At the individual algorithm level, AdaBoost & IG2 is the most-frequent worst combination, being the worst for 4 hard data sets.
- The worst combinations for 11 (73.3%) of the 15 hard data sets involve the use of feature selection methods; the worst combinations for 1 (6.7%) do not involve the use of feature selection methods; for the remaining 3 (20%), multiple combinations obtained the worst AUC, some of which involve feature selection methods and some do not.
- Focusing on the feature selection methods, we see that CFS is used in the worst combinations for 9 hard data sets (60%), IG2 is used for 5 hard data sets (33.3%), RLF1 and RLF2 and NO (no attribute selection) are used for 4 hard data sets (26.7%), and IG1 is used for 3 hard data sets (20%).
- Focusing on classification algorithms, we see that J48 and AdaBoost and SVM-PN appear in the worst combinations of 4 hard data sets (26.7%), and the others each appears in 1 (6.7 %).

So, for the 15 hard data sets, J48 and AdaBoost and SVM-PN are the most-frequent worst classification algorithms, CFS is the most-frequent worst feature selection methods, and AdaBoost & IG2 is the most-frequent worst combinations.

**Table 7**

Worst classification algorithm obtaining the lowest AUC for hard data sets.

| Data set | #Instance | #Attribute | #SelAttr | Classifier | FS method | Min48AUC |
|---|---|---|---|---|---|---|
| post-operative | 90 | 9 | 6 | IBK | RLF2 | 0.279 |
| planning | 182 | 13 | 13 | LOG | NO | 0.345 |
| heart-switzerland | 123 | 13 | 2 | RF | CFS | 0.454 |
| dresses_attribute_sales | 500 | 14 | 9 | J48 | CFS | 0.458 |
| dresses_attribute_sales | 500 | 14 | 9 | J48 | IG2 | 0.458 |
| heart-va | 200 | 13 | 9 | AdaBoost | IG2 | 0.484 |
| haberman-survival | 306 | 4 | 3 | J48 | RLF2 | 0.489 |
| statlog-australian-credit | 690 | 15 | 3 | SVM-PN | CFS | 0.5 |
| congressional-voting | 435 | 17 | 4 | SVM-RBF | CFS | 0.509 |
| congressional-voting | 435 | 17 | 4 | SVM-PN | CFS | 0.509 |
| pittsburg-bridges-span | 92 | 8 | 8 | AdaBoost | NO | 0.545 |
| pittsburg-bridges-span | 92 | 8 | 5 | AdaBoost | CFS | 0.545 |
| pittsburg-bridges-span | 92 | 8 | 7 | AdaBoost | IG1 | 0.545 |
| pittsburg-bridges-span | 92 | 8 | 6 | AdaBoost | IG2 | 0.545 |
| contrac | 1473 | 10 | 10 | AdaBoost | NO | 0.549 |
| contrac | 1473 | 10 | 4 | AdaBoost | CFS | 0.549 |
| contrac | 1473 | 10 | 9 | AdaBoost | IG1 | 0.549 |
| contrac | 1473 | 10 | 7 | AdaBoost | IG2 | 0.549 |
| contrac | 1473 | 10 | 9 | AdaBoost | RLF1 | 0.549 |
| contrac | 1473 | 10 | 7 | AdaBoost | RLF2 | 0.549 |
| breast-cancer | 286 | 10 | 5 | SVM-PN | CFS | 0.583 |
| spect | 265 | 23 | 19 | SVM-PN | RLF1 | 0.607 |
| primary-tumor | 330 | 18 | 18 | AdaBoost | NO | 0.609 |
| primary-tumor | 330 | 18 | 10 | AdaBoost | CFS | 0.609 |
| primary-tumor | 330 | 18 | 15 | AdaBoost | IG1 | 0.609 |
| primary-tumor | 330 | 18 | 12 | AdaBoost | IG2 | 0.609 |
| primary-tumor | 330 | 18 | 15 | AdaBoost | RLF1 | 0.609 |
| primary-tumor | 330 | 18 | 12 | AdaBoost | RLF2 | 0.609 |
| statlog-german-credit | 1000 | 25 | 21 | J48 | RLF1 | 0.653 |
| titanic | 2201 | 4 | 2 | J48 | CFS | 0.688 |
| titanic | 2201 | 4 | 2 | J48(-U-A) | CFS | 0.688 |

## 5.3 Maximum and Minimum AUC by All Combinations for the Hard/Easy Data Sets

For each data set, let min48AUC be defined similarly to max48AUC, and let span48AUC denote max48AUC – min48AUC.

Fig. 1 presents max48AUC and min48AUC for the hard data sets. We note that maximum span48AUC, minimum span48AUC, and average span48AUC are 0.257, 0.067 and 0.165 respectively. So the choice of classification and feature selection methods often has big impact on the classification accuracy for the hard data sets.

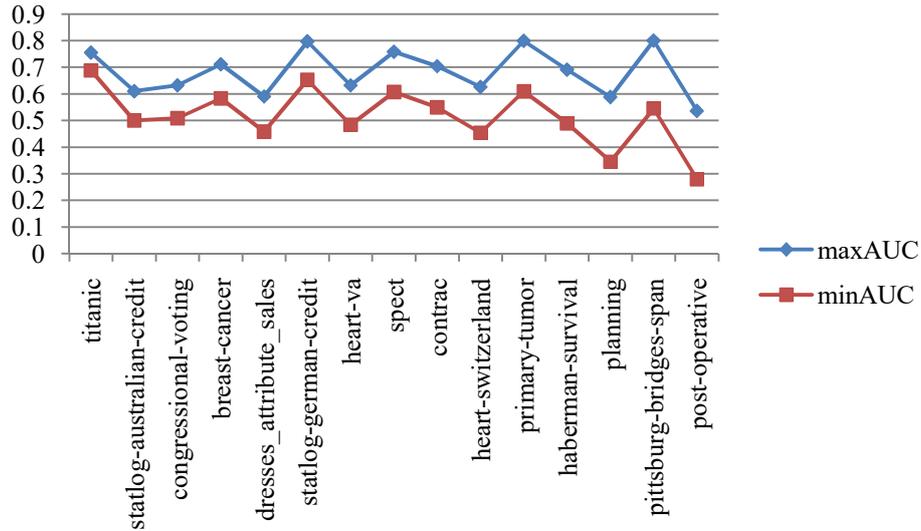

**Fig. 1.** Max48AUC and min48AUC for the 15 hard data sets.

Fig. 2 presents max48AUC and min48AUC for all easy data sets. We note that maximum span48AUC, minimum span48AUC, and average span48AUC are 0.77, 0 and 0.287 respectively. Moreover, span48AUC is greater than 0.3 for 48.2% of the easy data sets. While several easy data sets can be classified well by all of the 48 combinations, for nearly half of the easy data sets the difference in classification performance by different combinations is large.

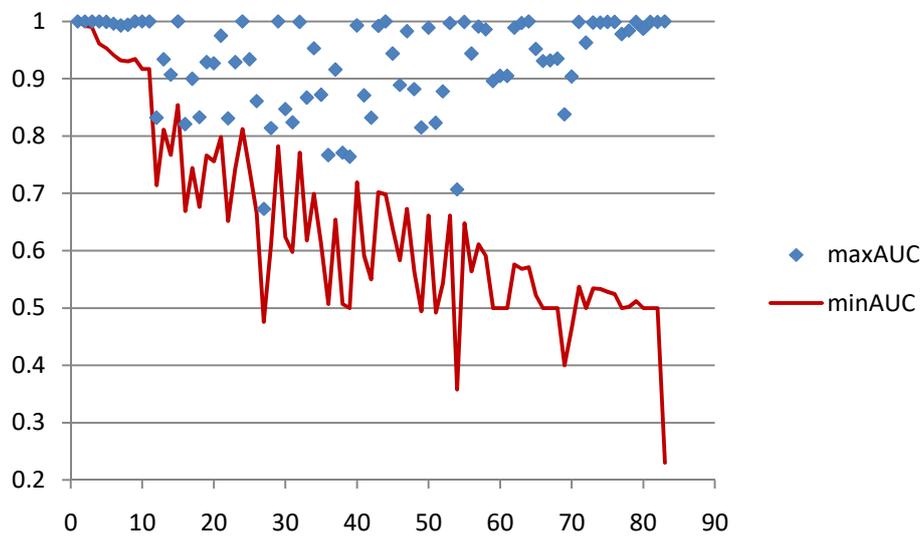

**Fig. 2.** Max48AUC and min48AUC for the easy data sets.

### 5.4 Average AUC and Average F-Measure for the Hard/Easy Data Sets

Fig. 3 shows the average AUC and average F-Measure for the eight classification algorithms on (1) the 15 hard data sets (upper panel) and (2) all of the easy data sets (lower panel). We observe that, based on average AUC, RF is the best for the hard data sets, and RF is also the best for the easy data sets.

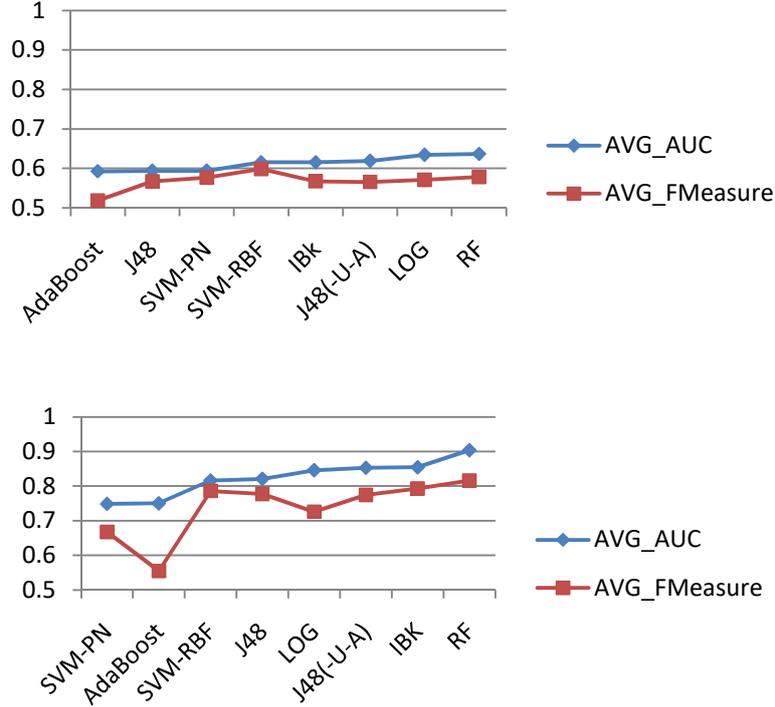

**Fig. 3.** Average AUC and average F-Measure for the eight classification algorithms on the 15 hard data sets (upper panel) and the easy data sets (lower panel).

The slope of the curve in the upper panel is gentle (almost flat): the maximum average AUC is 0.637, the minimum average AUC is 0.592, and their difference is 0.045; the maximum average F-Measure is 0.578, the minimum average F-Measure is 0.518, and their difference is 0.06. The slope of the curve in the lower panel is fairly steep: the maximum average AUC is 0.904, the minimum average AUC is 0.749, and their difference is 0.155; the maximum average F-Measure is 0.816, the minimum average F-Measure is 0.668, and their difference is 0.148. In summary, the difference among the performance of the classification algorithms for the 15 hard data sets is fairly small; in contrast, the difference for the easy data sets is fairly large.

We note that RF is the best for both easy and hard data sets, which is in strong agreement with the ranking of algorithms provided by [1]. However, Fig. 3 shows that SVM-PN is the last one in the rank for easy data sets, which is very different from the ranking give by [1] (which found SVM to be the second best classification algorithm). There are at least three potential reasons for the disagreement: We used AUC and F-Measure whereas [1] used accuracy. (2) In our study we excluded a number of data sets for which [1] reported high classification accuracies (by any of the classification algorithms) and we included some data sets not studied in [1]. (3) In our ranking, we separated data sets into a hard pool and an easy pool, whereas [1] considered all data sets in one pool.

Table 8 shows the average AUC and average F-Measure for the six feature selection methods on the hard data sets (left table) and the easy data sets (right table). For the hard data sets, the IG1 methods is the best; for the easy data sets, RLF1 is the best, followed by IG1 and NO (no attribute selection). We note that the average AUC and average F-Measure of IG1 and NO (no attribute selection) are all just 0.001 below those of RLF1.

More specifically, for the hard data sets (Table 8, left), the maximum average AUC is 0.62 and the minimum average AUC is 0.605, and the difference is just 0.015; the maximum average F-Measure is 0.576, the minimum average F-Measure is 0.561, and the difference is just 0.015. For the easy data sets (Table 8, right), the maximum average AUC is 0.831, the minimum average AUC is 0.81, and the difference is 0.021; the maximum average F-Measure is 0.743, the minimum average F-Measure is 0.725, and the difference is 0.018. The above suggests that there is little difference on the classification performance whether feature selection methods are used, or which feature selection methods are used, based on average performance. We must note that the above statement is based on average performance over a large number of data sets. As noted above, the exclusion of a fairly large number of data sets with known high classification accuracy may also contribute to the above findings.

**Table 8**

Average AUC and average F-Measure for the six feature selection methods on the hard data sets (left) and the easy data sets (right).

| Method | AVG_AUC | AVG_FMeasure |
|---|---|---|
| RLF2 | 0.605 | 0.561 |
| CFS | 0.605 | 0.561 |
| RLF1 | 0.612 | 0.567 |
| IG2 | 0.615 | 0.568 |
| NO | 0.618 | 0.575 |
| IG1 | 0.62 | 0.576 |

| Method | AVG_AUC | AVG_FMeasure |
|---|---|---|
| CFS | 0.81 | 0.725 |
| IG2 | 0.821 | 0.732 |
| RLF2 | 0.823 | 0.737 |
| NO | 0.83 | 0.742 |
| IG1 | 0.83 | 0.742 |
| RLF1 | 0.831 | 0.743 |

### 5.5 Summary of Classifier Rankings on Hard Data Sets and on Easy Data Sets

Tables 9 and 10 summarize the rankings of the 8 classification algorithms for the hard and easy data sets respectively. Each gives two classifier rankings, one based on the number of data sets for which the classifier is the best, and the other based on average AUC. #DatasetsBest is the number of data sets for which a given algorithm obtained the maximum AUC. There are several classifiers obtaining the maximum AUC for some easy data sets, so the sum of #DatasetsBest in Table 10 (left panel) is larger than the number of easy data sets.

**Table 9**

Classifier rankings for hard data sets: based on number of data sets an algorithm is the best (left), and based average AUC (right).

| Classifier | #DatasetsBest |
|---|---|
| RF | 7 |
| LOG | 4 |
| AdaBoost | 2 |
| SVM- RBF | 2 |
| SVM- PN | 0 |
| IBK | 0 |

| Classifier | AVG_AUC |
|---|---|
| RF | 0.637 |
| LOG | 0.634 |
| J48(-U-A) | 0.619 |
| IBk | 0.615 |
| SVM-RBF | 0.615 |
| SVM-PN | 0.594 |

| J48      | 0 | J48      | 0.594 |
| J48(-U-A)| 0 | AdaBoost | 0.592 |

Table 10

Classifier rankings for easy data sets: based on number of data sets an algorithm is the best (left), and based average AUC (right).

| Classifier | #DatasetsBest | Classifier | AVG_AUC |
|---|---|---|---|
| RF | 59 | RF | 0.904 |
| LOG | 22 | IBk | 0.854 |
| SVM-RBF | 11 | J48(-U-A) | 0.853 |
| J48(-U-A) | 10 | LOG | 0.846 |
| IBK | 9 | J48 | 0.821 |
| AdaBoost | 9 | SVM-RBF | 0.816 |
| J48 | 7 | AdaBoost | 0.75 |
| SVM-PN | 7 | SVM-PN | 0.749 |

# 6 Conclusions

This paper reported a systematic evaluation of classification performance by representative state-of-the-art classification algorithms and feature selection methods on 129 data sets from UCI. It identified a list of benchmark data sets that are hard for representative classification and feature selection methods. It ranked the classification algorithms based on their performance on the hard data sets, and on their performance on the easy data sets. It also compared the effectiveness of feature selection methods. To the best of our knowledge, this study is the first to give a list of hard benchmark data sets in the machine learning literature, and to rank classification algorithms by considering their performance on hard data sets. This list of hard benchmark data sets can be useful for motivating the development of new classification and feature selection algorithms, and for use in the evaluation of such algorithms.

**Acknowledgements**

This work was partly supported by the National Natural Science Foundation of China (Grant No. 61872231) and the Shanghai Natural Science Foundation of China (Grant No. 14ZR1419200).

The authors wish to thank Dr Huan Liu and Dr Lei Yu for encouraging us to carry out this study. Part of the first author's work was done while visiting the Data Mining Research Lab at Wright State University.

**Appendices**

Below, Table X gives the run time for the wrapper method on several data sets, showing that the method is very time consuming.

Table XI lists the 31 data sets for which our experiments with the 48 classification algorithm and feature selection method combinations were not completed due to reasons such as "lack of memory", "taking too much time" (that is, for some of the combinations, the 10 folds cross validation took more than 120 hours), "abnormal program termination" etc. For each of the 31 data sets computation for at least one of the 48 combinations was completed. From the partial results of the finished experiments, we get the maximum AUC and maximum F-Measure for these data sets. Based on the partial results we are quite certain that the 31 data sets all belong to the easy data set category except the first one. We are not sure whether the first data set is a hard data set or not, because its maximum AUC and maximum F-Measure are all less than 0.8 based on the partial results.

**Table X**

Run time for the wrapper method on several data sets, on a laptop with Intel 2.3GHz processor, 4GB RAM, and 64-bit operating system.

| Data set | Instance | Attribute | Classifier | Runtime(seconds) |
|---|---|---|---|---|
| pima | 768 | 9 | SVM-PN | 193623 |
| crx | 690 | 16 | SVM-PN | 105864 |
| anneal | 798 | 39 | RF | 19652 |

**Table XI**

Data sets for which experiments were not complete; the MaxAUC and MaxFMeasure were based on the finished experiments.

| Data set | #Instance | #Attribute | MaxAUC | MaxFMeasure |
|---|---|---|---|---|
| diabetic_data | 101766 | 50 | 0.662 | 0.537 |
| actrecog1 | 162499 | 5 | 1 | 1 |
| actrecog2 | 137730 | 5 | 1 | 1 |
| actrecog3 | 102339 | 5 | 1 | 1 |
| actrecog4 | 122199 | 5 | 1 | 1 |
| actrecog5 | 159999 | 5 | 1 | 1 |
| actrecog6 | 140669 | 5 | 1 | 1 |
| actrecog7 | 162999 | 5 | 1 | 1 |
| actrecog8 | 137794 | 5 | 1 | 1 |
| actrecog9 | 163739 | 5 | 1 | 1 |
| actrecog10 | 126799 | 5 | 1 | 1 |
| actrecog11 | 104449 | 5 | 1 | 1 |
| actrecog12 | 114700 | 5 | 1 | 1 |
| actrecog13 | 67649 | 5 | 1 | 1 |
| actrecog14 | 116099 | 5 | 1 | 1 |
| actrecog15 | 103499 | 5 | 1 | 1 |
| har-puc-rio | 165633 | 19 | 0.999 | 0.987 |
| jsbach_chorals_harmony | 5665 | 17 | 0.982 | 0.762 |
| mhealth_subject1 | 161280 | 24 | 0.996 | 0.974 |
| mhealth_subject2 | 130561 | 24 | 0.996 | 0.97 |
| mhealth_subject3 | 122112 | 24 | 0.997 | 0.972 |

| mhealth_subject4 | 116736 | 24 | 0.996 | 0.97 |
| mhealth_subject5 | 119808 | 24 | 0.996 | 0.969 |
| mhealth_subject6 | 98304 | 24 | 0.997 | 0.971 |
| mhealth_subject7 | 104448 | 24 | 0.996 | 0.963 |
| mhealth_subject8 | 129024 | 24 | 0.997 | 0.973 |
| mhealth_subject9 | 135168 | 24 | 0.994 | 0.961 |
| mhealth_subject10 | 98304 | 24 | 0.998 | 0.971 |
| plant-shape | 1600 | 65 | 0.978 | 0.64 |
| sensorless_drive_diagnosis | 58509 | 49 | 1 | 0.999 |
| wle | 39242 | 159 | 1 | 1 |